\documentclass{article}
\usepackage{ijcai22}

\usepackage{times}

\usepackage{soul}
\usepackage{url}
\usepackage[hidelinks]{hyperref}
\usepackage[utf8]{inputenc}
\usepackage[small]{caption}
\usepackage{graphicx}
\usepackage{amsmath}
\usepackage{booktabs}
\usepackage{color}

\usepackage{multirow}
\usepackage{array}
\usepackage{makecell}

\title{A Survey of Visual Sensory Anomaly Detection}

\author{
Xi Jiang$^{1}$\footnote{Authors contributed equally.}\and
Guoyang Xie$^{1, 2}$\footnotemark[1]\and
Jinbao Wang$^{1}$\footnotemark[1]\and
Yong Liu$^{3}$\and
Chengjie Wang$^{3}$\and
Feng Zheng$^1$\footnote{Corresponding Author}\And
Yaochu Jin$^{4, 2}$\footnotemark[2]\\
\affiliations
$^1$Southern University on Science and Technology, VIP Lab\\
$^2$University of Surrey, NICE Group\\
$^3$Tencent, Youtu Lab\\
$^4$Bielefeld University, Faculty of Technology\\
\emails
jiangx2020@mail.sustech.edu.cn,
\{choasliu, jasoncjwang\}@tencent.com,\\
\{wangjb, zhengf\}@sustech.edu.cn,
\{guoyang.xie, yaochu.jin\}@surrey.ac.uk
}

\begin{document}

\maketitle

\begin{abstract}
Visual sensory anomaly detection (AD) is an essential problem in computer vision, which is gaining momentum recently thanks to the development of AI for good. Compared with semantic anomaly detection which detects anomaly at the label level (semantic shift), visual sensory AD detects the abnormal part of the sample (covariate shift). However, no thorough review has been provided to summarize this area for the computer vision community. In this survey, we are the first one to provide a comprehensive review of visual sensory AD and category into three levels according to the form of anomalies. Furthermore, we classify each kind of anomaly according to the level of supervision. Finally, we summarize the challenges and provide open directions for this community.~All resources are available at~\href{https://github.com/M-3LAB/awesome-visual-sensory-anomaly-detection}{https://github.com/M-3LAB/awesome-visual-sensory-anomaly-detection}.

\end{abstract}

\begin{figure*}
    \centering
    \includegraphics[width=\linewidth]{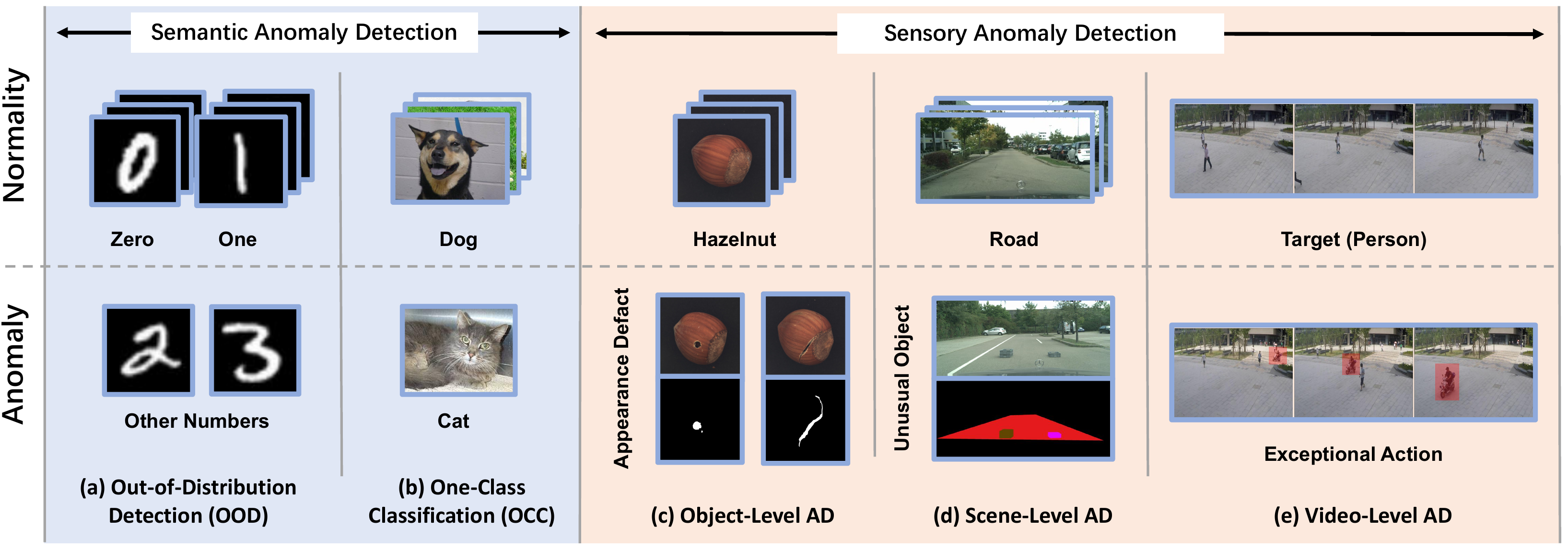}
    \caption{Exemplar problem settings for tasks under anomaly detection framework. (a) In out-of-distribution detection, multi-modal semantic anomaly detection uses supervised class labels. (b) In one-class classification, normality revolves around single-class data and anomaly is images with the semantic shift. (c) In object-level AD, normality is healthy objects(industrial products or organs, etc.) and the anomaly is a defect that occurs on it. (d) In scene-level AD, normality is a kind of conventional scene and anomaly is some novelty objects. (e) Compared with the previous anomaly detection on images, the video is the carrier of video-level anomaly detection, and the anomaly is manifested as abnormal behavior or event in videos. 
    Dataset: (a) MNIST~\protect\cite{mnist}, (b) Dogs VS. Cats~\protect\cite{dogs}, (c) MVTec AD~\protect\cite{16}, (d) Lost and Found~\protect\cite{lostandfound}, (e)    ShanghaiTech~\protect\cite{shanghaitech}.}
    \label{fig:1}
\end{figure*}

\section{Introduction}

Anomaly means a deviation from a rule or from what is regarded as standard, normal, or expected. 
The occurrence of anomalies is often unpredictable and the baneful influence is difficult to estimate without distinguishing them.
Therefore anomaly detection (AD) has been identified as a significant research field. 
In this review, we emphasize a challenging task, real-world AD in visual materials, in which anomaly can be a small defect in an industrial product, an unusual object in a driving scene, or an exceptional action in a video.


\begin{table*}[th]\tiny
\setlength{\abovecaptionskip}{2pt}
\caption{Summary of visual sensory anomaly detection.}
\renewcommand{\arraystretch}{1.3}
\resizebox{\textwidth}{!}{
\begin{tabular}{c|c|c|c}
\hline
    \textbf{AD}        &   \textbf{Taxonomy}    & \textbf{Learning Paradigm}   &  \textbf{Methods} \\\hline
    \multirow{7}{*}{\makecell[c]{Image \\ \textcolor{blue}{(Sec.~\ref{sec:object} \& \ref{sec:scene})}}} & \multirow{5}{*}{\makecell[c]{Object Level \\ \textcolor{blue}{(Sec.~\ref{sec:object})}} } & \makecell[c]{Unsupervised \\ \textcolor{blue}{(Sec.~\ref{sec:object_unsup})}}       & \multicolumn{1}{m{8cm}}{ \textit{Self-supervised:} \cite{2} \cite{5}  \textit{Pre-trained adaption:} \cite{3}  \cite{11} \textit{Image reconstruction:} \cite{10}  \cite{17} \textit{Feature modeling:} \cite{6} \cite{9} } \\\cline{3-4}
                             &                               & \makecell[c]{Weakly Supervised \\ \textcolor{blue}{(Sec.~\ref{sec:obect_weak})}}   & \multicolumn{1}{m{8cm}}{ \textit{Incomplete/semi-supervised:} \cite{12} \cite{14}  \cite{25} \textit{Inexact supervised:} \cite{15} }\\\cline{3-4}
                             &                               & \makecell[c]{Supervised   \textcolor{blue}{(Sec.~\ref{sec:object_super})}}         & \multicolumn{1}{m{8cm}}{\textit{Concrete structural defect:} \cite{18}   \textit{Fabric defect:} \cite{20} \cite{26} }\\\cline{2-4}
                             & \multirow{2}{*}{\makecell[c]{Scene Level \\ \textcolor{blue}{(Sec.~\ref{sec:scene})}}}  & \makecell[c]{Weakly Supervised \\ \textcolor{blue}{(Sec.~\ref{sec:scene_weak})}} &\multicolumn{1}{m{8cm}}{ \cite{38} \cite{36} \cite{31} \cite{32}  }  \\\cline{3-4}
                             &                               & \makecell[c]{Supervised  \textcolor{blue}{(Sec.~\ref{sec:scene_super})}}       & \multicolumn{1}{m{8cm}}{ \cite{41} \cite{42} \cite{43} }\\\hline
\multirow{3}{*}{\makecell[c]{Video \\ \textcolor{blue}{(Sec.~\ref{sec:event})}}} & \multirow{3}{*}{\makecell[c]{Event Level \\ \textcolor{blue}{(Sec.~\ref{sec:event})}}}  & \makecell[c]{Unsupervised \\ \textcolor{blue}{(Sec.~\ref{sec:event_unsup})}}       & \multicolumn{1}{m{8cm}}{ \cite{Barz2019DetectingRO} \cite{Chang2020ClusteringDD} \cite{Georgescu2021ABF} \cite{Gong2019MemorizingNT}   \cite{Li2020QuantifyingAD} \cite{Luo2021FutureFP}    }\\\cline{3-4}
                             &                               & \makecell[c]{Weakly Supervised \\ \textcolor{blue}{(Sec.~\ref{sec:event_weak})}}  & \multicolumn{1}{m{8cm}}{ \cite{Lv2021LocalizingAF} \cite{PurwantoDanceWS}  \cite{Tian2021WeaklysupervisedVA} \cite{Wu2021WeaklySupervisedSA}  \cite{Zaheer2020CLAWSCA}  } \\\cline{3-4}
                             &                               & \makecell[c]{Supervised   \textcolor{blue}{(Sec.~\ref{sec:event_super})}}         & \multicolumn{1}{m{8cm}}{ \cite{Leyva2017VideoAD} \cite{Li2022VariationalAB} \cite{Liu2018FutureFP}   \cite{Lu2018FastAE}  }  \\\hline
\end{tabular}
}\label{tab:summary}
\end{table*}

\textbf{Related Reviews and Surveys} Out-of-distribution (OOD) detection is closely related to AD, because the abnormal part in the sample can be regarded as out of the normal distribution, and an OOD sample can also be regarded as an anomaly in the sample set. 
As an important research area, there have been several surveys~\cite{survey1,survey3,survey4}, focusing specifically on this macro domain and regarded AD as a very general notion. 
Some typical research~\cite{maziarka2021oneflow} also regarded narrow sense OOD detection and AD as the same research direction. 
However, narrow sense OOD is a sample level exception, and recent viewpoint~\cite{yang2021generalized} provided that AD contains covariate shift detection which detects special sensory anomalies that do not fall within the narrow definition of OOD. 
Another related topic is one-class classification (OCC). Different from OOD, OCC concentrates on single-class (uni-modal) data. But both of them rely on supervised class labels for each class, in other words, act on semantic shift. 

The narrow sense OOD tends to set up label exceptions on existing categorical data sets, which is far from the real world. 
More and more of the most cutting-edge hot work emphasize visual sensory AD but not detecting semantic outliers. However, none or only some of the known surveys specifically summarize these works. 
Instead, we specifically investigated visual sensory AD with covariate shift, in which only an area of the visual sample may be abnormal while in the label space the whole is subject to the identical distribution. 

Most existing surveys~\cite{survey4,survey6} classify former researches in terms of solutions in the traditional way which aims at only one type of anomaly and does not even look at the essential characteristics of the task of supervision patterns. 
An integrated treatment of various real-world anomalies detection with deep learning methods is still missing.

\textbf{Scope of this Survey} In this survey, we try to categorize different researches based on the kinds of anomalies in recent years. According to the scenarios and forms of anomalies, the anomalies in existing studies can be divided into three categories: object-level, scene-level, and event-level. 
In object-level AD, an anomaly exists in a small part of an object, which may be an industrial defect, a medical disease feature, etc. In scene-level AD, an anomaly can be an object in an image scene(e.g. obstacle on-road). Event-level AD is set for the video data, in which anomaly is an action or event. Fig.\ref{fig:1} illustrates the difference between semantic AD and sensory AD and the settings of each category. 

We further classify research methods according to the learning manner. Because there is no unified and accurate definition of AD supervision, we developed a three-category standard, including unsupervised, weakly supervised, and fully supervised learning. 
In unsupervised learning, AD does not use any annotation information and only contains normal data. In weakly supervised learning, labels for normal or abnormal data are available, but are incomplete, inexact, or inaccurate ~\cite{zhou2018brief}. Supervised learning occurs when both normal and abnormal are marked. For the last category, some scholars ~\cite{survey4,survey6} include the fully supervised method in AD while others ~\cite{survey1} do not. By investigating existing supervised methods, we found that supervised AD should not be confused with naive detection and segmentation tasks, and classifying it is helpful for further research. 

Finally, potential future lines of research are provided practically to address the issues of current methods. In summary, the framework of this survey is shown in Table~\ref{tab:summary}. The taxonomy of this survey consists of three aspects, namely object-leave AD, scene-level AD, and event-level AD.


Our main contributions are summarized as follows:
\begin{itemize}
    \item To the best of our knowledge, it is the first work to provide a comprehensive review for visual sensory AD, which pays more attention to realistic vision tasks.  
    \item We are the first one to review the current visual AD studies categorized by anomaly level. Our work can help the community better understand the essence of anomaly. 
    \item The proposed work is the first one to summarize the main issues and potential challenges in visual sensory AD, which outlines the underlying research directions for future works. 
\end{itemize}


\section{Object-level AD} \label{sec:object}
Object-level AD is the most immediate use of AD in the real world. It can be widely used in industrial production, medical diagnosis, product maintenance, and other aspects. 
Some surveys~\cite{survey6} divide visual AD into two different research themes: supervised and unsupervised but restrict the line of sight to unsupervised image anomalies. 
We can obtain examples of abnormal samples in the real world. Though difficult to obtain, relying on this information can help deep learning models better distinguish anomalies. 
We summarize object-level AD methods into unsupervised learning with no anomaly data available in training, weakly supervised learning with incomplete or inexact anomaly data in training, and supervised learning with more sufficient anomaly data.

\subsection{Unsupervised Method}\label{sec:object_unsup} 
\begin{figure*}[ht]
    \centering
    \includegraphics[width=\linewidth]{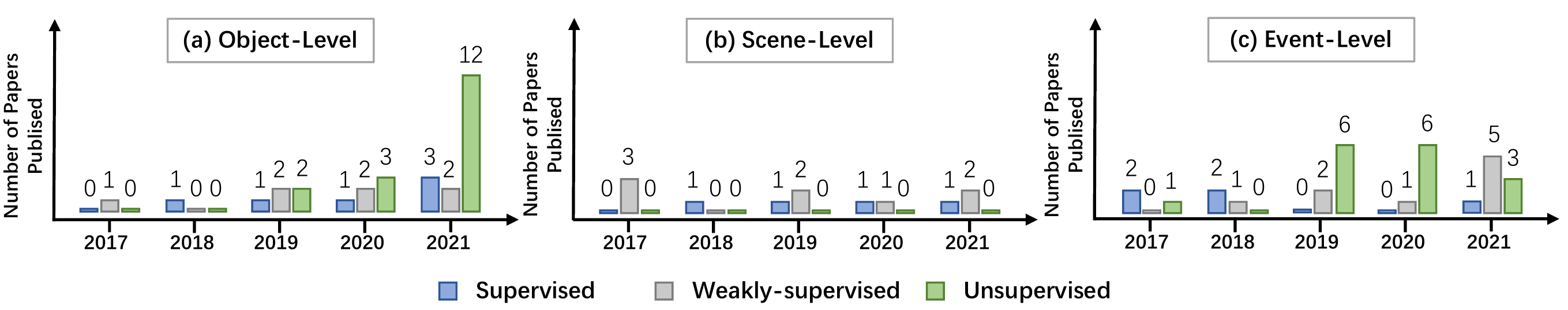}
    \caption{Trend of each learning manner in visual sensory anomaly detection in the past five years. The papers in statistics are from the top conferences and journals in the area of computer vision and machine learning. }
    \label{fig:trend_diff_learning}
\end{figure*}

Unsupervised object-level AD is one of the most interesting topics in this survey. Thanks to the open data set MVTec AD ~\cite{16}, studies have poured out in recent years as shown in figure~\ref{fig:trend_diff_learning}.
Since there is no abnormal data as a reference, the deep learning model needs various algorithms to familiarize it with the normal sample information to distinguish abnormal samples and delineate abnormal regions. 

\textbf{Self-supervised Method} One of the most efficient methods is self-supervised. By introducing a self-designed agent task, the model can better extract and fit the features of normal samples and improve the sensitivity to anomalies. 
Naively applying existing self-supervised tasks, such as rotation prediction or contrastive learning, is sub-optimal for detecting local defects~\cite{2}. So ~\citeauthor{2} innovatively design a proxy task for learning representation and name the augmented method as CutPaste. CutPaste simulates an abnormal sample by clipping a patch of a normal image and pasting it back at a random location. Despite the significant gap between the synthetic anomaly and the real defect, learning on CutPaste to find irregularity generalizes well on unseen anomalies. 
~\citeauthor{5}~[\citeyear{5}] employ transformations such as horizontal flip, shift, rotation, and gray-scale transformation after a multi-scale generative model to enhance the representation. 

\textbf{Pre-trained Adaption} Pre-trained adaption is another promising direction to obtain better features. We can pre-train the model on a labeled data set, which enhances the ability to learn deep representations compared to unsupervised learning. 
Naive transferring pre-trained representations from typically supervised learning to the one-class classification (OCC) task often results in catastrophic collapse (feature deterioration)~\cite{3}. ~\citeauthor{3}~[\citeyear{3}] find that feature adaptation with constant-duration early stopping already achieves better performance and elastic regularization inspired by continual learning is also of great use. 
~\citeauthor{11}~[\citeyear{11}] pre-train a teacher model on a large dataset of natural images, then uses a student model to imitate the teacher's output in unsupervised AD training. Anomalies are detected by contrasting the outputs of the teacher network and that of the student network which is difficult to generalize from anomaly-free training data. 

\textbf{Image Reconstruction}
Image reconstruction requires the model to reconstruct abnormal images into normal ones. 
~\citeauthor{10}~[\citeyear{10}] use multi-scale horizontal and vertical stripes to mask normal images and trains GAN to recover the unseen regions using context. 
But local defects added to a normal image can deteriorate the whole reconstruction~\cite{7}. By iteratively updating the input of the autoencoder and adding prior knowledge about the expected anomalies via regularization terms, higher quality images are produced than ones generated by classic reconstructions. 

\textbf{Feature Modeling}
Feature modeling detects OOD in latent space via extracting feature maps and establishing a memory model of normal samples. This method is relatively effective for Semantic AD, but for Sensory AD, its application is more difficult, because the abnormal features are not obvious. 
~\citeauthor{9}~[\citeyear{9}] leverage the descriptiveness of multi-scale features extracted in the defect detection model to estimate their density using normalizing flows. By calculating likelihoods from the generated well-defined distribution, the characteristics of the abnormal samples will be out of distribution and hence have lower likelihoods than normal images. 
Storing feature distribution in memory is another fancy modeling way which is first used in video-level AD by ~\citeauthor{Gong2019MemorizingNT}~[\citeyear{Gong2019MemorizingNT}]. 
The granularity of division on feature maps is closely related to the reconstruction capability of the model for both normal and abnormal samples~\cite{6}. So a multi-scale block-wise memory module is embedded into an autoencoder network by ~\citeauthor{6} which processes reconstruction of an image as a divide-and-assemble procedure. 

Across all approaches to unsupervised learning, the challenge from an information perspective is twofold:
1) How to use existing knowledge (i.e., normal data) to distinguish anomalies? In previous work, the detection system remembers the normal data distribution or can reconstruct abnormal images into normal images through modeling or reconstruction. Compared with semantic AD, sensory AD often requires higher system capability. Better ways of memorizing or modeling distributions, faster detection, better utilization of normal data areas, and so on are all aspects of practical significance for future work. 
2) How to use the additional knowledge to guide the model to recognize exceptions? The extra knowledge here can be either an extra data set or a human guide on how to identify anomalies. Past work has helped improve the model's feature extraction capabilities by introducing additional data and improving the model's perception by using an artificial anomaly sample (albeit different from the ones being detected in the test set). Due to the small size and uneven distribution of AD data sets, this additional knowledge is often very helpful for identification. As more and more object-level anomaly datasets become available, how to transfer detection capabilities among different data distributions is also an interesting work in the future, because it is intuitively possible that these other types of anomaly data are fundamentally useful. 
\subsection{Weakly Supervised Method}\label{sec:obect_weak}
In reality, there are always real-world exceptions that can be used to aid training, but they can often be incomplete or imprecise. Weakly supervised methods are designed to adequately use the information contained in these negative samples. In different scenarios, these methods combine the advantages of the precision found in supervised methods and the substantial reduced need for large amounts of data in unsupervised methods. 

The incomplete supervised setting, also known as semi-supervised, only uses a handful of labeled anomalous data in training. 
~\citeauthor{12}~[\citeyear{12}] observe that when the model is trained on such imbalanced data, the loss values over a normal sample decreases, while the loss values on an anomalous sample fluctuate. This kind of interesting trend can be used as a feature to identify anomalous data via the introduced reinforcement learning-based meta-algorithm, neural batch sampler, though it needs several training procedures to get the loss profiles which is the basis for judging exceptions. 
FCDD ~\cite{30} is a fully convolutional one-class classification method where the output features serve as an anomaly heatmap. To concentrate nominal samples in feature space and map anomalies away, FCDD uses a Hypersphere Classifier and modifies with pseudo-Huber loss, utilizing the sum of output feature map as the anomaly score. They find that even using a few examples as the corpus of labeled anomalies performs exceptionally well. 
CAVGA~\cite{14} comprises a convolutional latent variable to preserve the spatial information between the input and latent variable. Then it learns to localize the anomaly using an attention map reflected through backpropagation to the last Conv layer of the encoder. In both unsupervised and weakly supervised settings, the attention-based loss helps the autoencoder focus on normal areas. 

CAVGA~\cite{14} is also an inexact supervised method. In this setting, the annotation of anomaly is not exact to the pixel-wise segmentation but can be a rough area or just an image-wise label. Fixed-Point GAN ~\cite{15} is inspired from image-to-image translation and applies domain translation into medical images which targets the healthy state. The introduced fixed-point translation allows the GAN to identify a minimal subset of pixels for domain translation. 

Weakly supervised method is more realistic for its more inclusive data setting, while domain adaptation is another interesting research in AD. 
In the real inspection system, distributional shift is widespread between training data and inference data due to distinct lighting conditions and changing specification of panels. To handle this problem, A-SVDD ~\cite{25} uses domain adaptation in unsupervised and semi-supervised settings to learn an incremental classifier based on the existing support vector data description. 

Weakly supervised learning has more information than unsupervised learning, and how to make the best use of these labeled data is the emphasis of research. 
In existing studies, these annotated anomaly data are used to help models emphasize on specific regions, learn additional features, or enhance mapping capabilities.
Although these directions are not fully developed, thanks to these label information, the weakly supervised system has stronger robustness compared with the unsupervised case, which makes the domain adaptation also become one of the research directions. 
Beyond that, another kind of weakly supervised learning, inaccurate learning, also called noisy label learning, has gone unnoticed in the field of AD, which is a case of more real-world application and more meaningful for future research.

\subsection{Supervised Method} \label{sec:object_super}
Supervised object-level AD is often required to detect multiple defect types in the industry or various illness from medical imaging, and this setup is similar to object detection and instance segmentation. These object-based methods are the basis of supervised AD, whose research progress is also promoting anomaly targeting methods. However, due to different tasks, special settings are required to deal with the challenges caused by various kinds of anomalies. 

Fabric and metal surface defects are common and readily available in the real world, therefore supervised methods can be applied with large-scale data sets. 
However, \citeauthor{20}~[\citeyear{20}] put forward that the texture shift and partial visual confusion are still challenging when directly applying the vanilla object detection method to the defect detection task. Texture shift describes the model generation ability challenge caused by different texture distribution, and partial visual confusion describes the challenge of similarity between partial defect boxes and complete defect boxes to a classifier. 
To detect structure defects that are more damaged and more difficult to identify compared with color defects, ~\citeauthor{26}~[\citeyear{26}] use a high resolution vision-based tactile sensor, which avoids the influence of dyeing patterns. Although the sense of touch is another mode of information, it is still displayed and processed in the form of images, which has good adaptability to neural network. 

Surface flaws on concrete structures are also a common anomaly in the real world. 
iDAAM~\cite{18} proposes that the concrete defects in the real-world issue are typically found in an overlapping manner. But most previous studies do not address it and are unable to apportion higher importance to the defective region from the healthy region. 
The author proposed interleaved fine-grained dense module to obtain robust discriminative features from similar-looking overlapping defect classes,  and a concurrent dual attention module to extract salient discriminative features from multiple scales and to address the part deformation and shape variation present in overlapping defect classes. 

Compared with object detection, supervised object-level AD faces more challenges, such as texture Shift, partial visual confusion and overlapping.
At the same time, the anomaly on the object is different from the specific object, each anomaly is different in size, shape and color, and its classification label is only a kind of artificial division, which is more rough compared with the label in object recognition. 
Therefore, future research can not only draw inspiration from areas such as object detection and medical images, but also combine with unsupervised and weakly supervised AD. 


\section{Scene-level AD} \label{sec:scene}

Scene-level AD is to detect anomaly in a scene of images, where the anomaly is usually an object. Compared with object-level images, scene-level images contain richer contents, broader perspectives, larger anomalies and more diverse types. 

\subsection{Weakly Supervised Method} \label{sec:scene_weak}
Classical object detection or semantic segmentation assume that all classes observed at test time have been seen during training. However, a more realistic scenario is that novelty objects of unknown classes can appear at test time.
Anomaly segmentation solves this problem of unmarked content in a scene segmentation task. Because normal categories are marked while abnormal categories are not, it is classified as weakly supervised learning. 

~\citeauthor{38}~[\citeyear{38}] have observed that the spurious labels will be produced in areas existing unexpected objects by encoders whose corresponding labels are absent from standard segmentation training datasets. Leveraging the reconstructive image from the semantic map, different regions of appearance can be lined out compared with the original image. 
Compared with the image reconstruction methods in unsupervised image-level AD that tries to re-synthesize normality, this similar approach at the scene level exploits the spurious semantics of the exception object and refactor semantically sound scenarios even if it does not make sense visually. 
The proposed discrepancy network not only outperforms the datasets depicting unexpected objects but also is generic to other tasks such as defencing adversarial attacks in segmentation. 

~\citeauthor{31}~[\citeyear{31}] mention that there are three possible outcomes when a segmentation network encounters anomalous objects. In the first situation, anomalies are segmented as other normal objects. Second, 
anomalies are divided into combinations of objects. In the last case, the exception is not detected and isolated. Using the segmentation uncertainty to improve over existing reconstruction methods, the proposed method outperforms previous approaches which used only one of the uncertainty and reconstruction methods for all scenarios. 
JSR-Net~\cite{32} fuses the pixel-level re-synthesis module with semantic segmentation network and joins the segmentation coupling module which couples the structural similarity error and information from known classes to produce an anomaly map. The advantages of this combination can prevent the segmentation network from being overconfident in similarity estimation, which often leads to the misclassification of small anomalous objects.

Similar to the object-level AD, the domain shift or distribution shift mentioned in Sec.~\ref{sec:obect_weak} is also a real-world problem. ~\citeauthor{36}~[\citeyear{36}] mention that because of the presence of novelty objects in the scene, the neural network needs to be updated with feedback loops for detected anomalies. Therefore, the author innovatively detects novel categories in semantic segmentation and retrieves their semantic similarity to enhance the domain shift capability of the model. 


Weak supervision is the main research direction of AD at the scene level. 
However, these kinds of method require normal instance labels to satisfy fundamental segmentation task, which is hard to meet for the expensive and time consuming annotation. Unsupervised methods have not been explored. 
Moreover, only under weak supervision, compared with object-level, the development direction of the method is closely centered on segmentation and reconstruction. 
Although this may be due to more complex semantic information at the scene level, feature modeling and pre-training adaptation may also contribute to performance.

\subsection{Supervised Method} \label{sec:scene_super}
Just as the small obstacle detection task can be represented as the semantic segmentation of the obstacle, the supervised scenario-level AD research usually only aims at the nature of the anomaly under specific circumstances, while the more general segmentation method can play a role when the anomaly label is sufficient.

MergeNet~\cite{41} tries to handle the challenge from limited availability of annotated data and proposes a multi-stage model architecture with RGBD image as input, which can be trained using 135 images to achieve better performance in comparison to the 1000 images used by baseline. 
Real time is also an important index in road obstacle detection. RFNet~\cite{42} employs the multi-scale information feature image extraction and fusion to exploit cross-model information efficiently and achieves 22 Hz inference speed at high resolution RGBD image. 
To solve the problem of missing the contour of remote micro-obstacles in monocular images, ~\citeauthor{43}~[\citeyear{43}] propose an obstacle perception method based on the knowledge of spatial perspective. Utilizing the pseudo distance and several novel obstacle-aware occlusion edge maps, the method can enclose tiny obstacles as much as possible. 


Generally speaking, the research on scene-level AD mainly focuses on road or autonomous driving scenarios, and AD in other scenarios is also an interesting task. For example, in indoor scenes, there will be more occlusion, so how to detect novel objects in more and more complex scenes remains a challenge.


\section{Event-level AD} \label{sec:event}

\subsection{Unsupervised Method} \label{sec:event_unsup}
The mainstream of unsupervised learning for video AD is to use auto-encoder or GAN to project the anomaly-free and anomaly ones into one common space. The next step is to adopt the unsupervised learning method, like clustering algorithm to detect the outliers~\cite{Chang2020ClusteringDD,Georgescu2021ABF,Gong2019MemorizingNT}. ~\citeauthor{Barz2019DetectingRO}~[\citeyear{Barz2019DetectingRO}] propose a uniform model to detect anomaly regions for time and space-varying measurements. In specific, the authors propose a "maximally divergent intervals" framework to detect isolated anomalous data points according to the high Kullback-Leibler. ~\citeauthor{Chang2020ClusteringDD}~[~\citeyear{Chang2020ClusteringDD}] propose a convolution neural network (CNN) based auto-encoder to capture spatial and temporal features, separately. In addition, the authors employ deep k-means cluster to find out the outliers. ~\citeauthor{Georgescu2021ABF}~[~\citeyear{Georgescu2021ABF}] apply GAN into the context of abnormal event detection. Furthermore, the authors focus on object detection, which could be more general and background agnostic. Finally, ~\citeauthor{Georgescu2021ABF}~[~\citeyear{Georgescu2021ABF}] offer regions-level and track-level annotations for the downstream tasks. ~\citeauthor{Gong2019MemorizingNT}~[~\citeyear{Gong2019MemorizingNT}] develop a new learning mechanism to mitigate the drawback of the auto-encoder generalization problem. Specifically, the encoder uses the most relevant memory items for reconstruction at the training phase so that the reconstruction could be more close to the anomaly-free ones. 
~\citeauthor{Li2020QuantifyingAD}~[~\citeyear{Li2020QuantifyingAD}] pay more attention to the anomalies in the crowded scenes. The authors of~\cite{Li2020QuantifyingAD} collect the invariant features from time, motion and fuse them to detect the anomaly ones. 
~\citeauthor{Luo2021FutureFP}~[~\citeyear{Luo2021FutureFP}] incorporate the meta-learning into the video AD problem, which helps the model converge faster in an unseen test domain. 

\subsection{Weakly Supervised Method} \label{sec:event_weak}
Classical semi-supervised video AD assumes that the anomaly data is not available in the training data set. ~\citeauthor{Lv2021LocalizingAF}~[~\citeyear{Lv2021LocalizingAF}] propose a high-order encoding model to simultaneously extract the semantic representations but also measure the time-varying features. In addition, the authors of~\cite{Lv2021LocalizingAF} use the segmentation task information as the auxiliary information to obtain the final anomaly scores. ~\citeauthor{PurwantoDanceWS}~[~\citeyear{PurwantoDanceWS}] employ a relation-aware CNN to extract features from video. And then the authors of~\cite{PurwantoDanceWS} incorporate the self-attention mechanism into conditional random field (CRF). ~\citeauthor{Tian2021WeaklysupervisedVA}~[~\citeyear{Tian2021WeaklysupervisedVA}] propose a robust temporal feature magnitude learning to recognize the positive instances, which can increase the robustness of multiple instance learning problems. ~\citeauthor{Wu2021WeaklySupervisedSA}~[~\citeyear{Wu2021WeaklySupervisedSA}] propose another weakly supervised learning, i.e., just coarsely annotate the abnormal event in the video sequence and their task is to finely localize the anomaly event in each video frame. ~\cite{Wu2021WeaklySupervisedSA} propose a dual-path based mutually-guided progress refinement framework to train this model. There are three contributions of ~\citeauthor{Zaheer2020CLAWSCA}~[~\citeyear{Zaheer2020CLAWSCA}]. Firstly, to reduce inter-batch correlation, the authors propose a random batch-based training procedure. Secondly, the author put forward a novel suppression mechanism to minimize the anomaly scores. Finally, the author proposes a clustering distance-based loss to mitigate the label noise.

\subsection{Supervised Method} \label{sec:event_super}
Supervised video detection assumes that the anomaly data are located in the training and test dataset and their label is provided in the training dataset~\cite{Liu2018FutureFP}. ~\citeauthor{Leyva2017VideoAD}~[~\citeyear{Leyva2017VideoAD}] propose a coarse-to-fine hand-crafted features extraction manner. In specific, the features are extracted from foreground occupancy and optical flows. The author compact the features by using Gaussian Mixture Models.  Shortly, ~\citeauthor{Lu2018FastAE}~[~\citeyear{Lu2018FastAE}] propose a fast  abnormal event detection with online learning. The method rates at a speed of 1,000-1,200 frames per second on average.  
Recently, ~\citeauthor{Li2022VariationalAB}~[~\citeyear{Li2022VariationalAB}] adopt a motion loss to make the features of generated videos more consistent with the input videos.

\section{Open Challenges}

Although visual sensory AD has achieved significant progress, challenges still exist due to the complexity of the task. Here, we provide several future directions for this field.
\begin{itemize}
    \item \textbf{Pretrained Model for Industrial Detection} Nowadays, all of the pretrained models deployed in industrial AD are using the ones trained in ImageNet~\cite{Russakovsky2015ImageNetLS}, like ResNet18~\cite{He2016IdentityMI}. However, the bias exists inevitably due to the domain shift problem. So it's necessary to provide pretrained models trained by visual sensory anomalies. Although the number of anomaly data is limited, ~\citeauthor{Cao2022TrainingVT}~[~\citeyear{Cao2022TrainingVT}] indicate it is still possible to train the pretrained model with a small amount of data.
    \item \textbf{Continual Learning} Most of the industrial anomaly data are presented in an assembly line. So the AD model should overcome the catastrophic phenomenon since all of the data presented order are online.
    \item \textbf{3D AD} Recently, 3D industrial AD data is proposed by ~\citeauthor{Bergmann2021TheM3}~[~\citeyear{Bergmann2021TheM3}]. However, there are still missing the specific baseline model for 3D AD. This data has proved that the 3D AD method could be superior to 2D AD due to the auxiliary depth information.
    \item \textbf{Scenes Category Limited} Various agent tasks are used in object-level AD and achieve powerful performance in self-supervised learning as in Sec.~\ref{sec:object_unsup}. Because of the rich semantic information in scenarios, it should be easier to design agent tasks for scene-level AD, but no such research has been done in previous articles. Future research can combine existing semantics to generate novel objects in the scene. 
\end{itemize}


\small
\bibliographystyle{named}
\bibliography{ijcai22}

\end{document}